\documentclass[conference]{IEEEtran}
\usepackage{cite}
\usepackage{graphicx}

\usepackage[utf8]{inputenc}
\usepackage{tikz,float}
\usepackage{tabularx}
\usepackage{xparse}
\usetikzlibrary{shapes,backgrounds,calc}

\usepackage{graphicx} 
\usepackage{caption}
\usepackage{subcaption}
\usepackage{amsmath}
\usepackage{amsfonts}
\usepackage[cal=rsfso,calscaled=.96]{mathalfa}
\usepackage{mathrsfs}  
\usepackage{txfonts} 
\usepackage{enumerate}
 
\usepackage{hyperref}  
\usepackage{amsmath}
\usepackage{algorithm}
\usepackage[noend]{algpseudocode}

\usepackage{times} 
\usepackage{amssymb,longtable,calc}
\usepackage{mathptmx}       
\usepackage{graphicx}
\usepackage[textwidth=2cm,colorinlistoftodos]{todonotes}
\usepackage{xspace}
\usepackage{latexsym}
\usepackage{amsmath,amssymb,graphicx,xspace}
\usepackage{times}   
\usepackage{multirow}
\usepackage{color,comment,rotating,url,xspace} 
\usepackage{tikz}
\usetikzlibrary{arrows,shadows,shapes,backgrounds,decorations,snakes,fit}
\usepackage{sidecap}
\usepackage{amsmath}
\usepackage{subcaption}

\usepackage{amsfonts}

 \newcommand{\bmxt}{\mathbf{x_t}}

\newcommand{\bmzt}{\mathbf{z_t}}

\newcommand{\bmhlt}{h_{t-1}} 

\newcommand{\bmyt}{\mathbf{y_t}}

\newcommand{\pem}{\widetilde{p}}

\usepackage[top=1in, bottom=1in, left=1.25in, right=1.25in]{geometry}

\usepackage{todonotes}
\usepackage{enumerate}

\makeatletter
\tikzset{circle split part fill/.style  args={#1,#2}{%
 alias=tmp@name, 
  postaction={%
    insert path={
     \pgfextra{%
     \pgfpointdiff{\pgfpointanchor{\pgf@node@name}{center}}%
                  {\pgfpointanchor{\pgf@node@name}{east}}%
     \pgfmathsetmacro\insiderad{\pgf@x}
      \fill[#1] (\pgf@node@name.base) ([xshift=-\pgflinewidth]\pgf@node@name.north) arc
                          (90:270:\insiderad-\pgflinewidth)--cycle;
      \fill[#2] (\pgf@node@name.base) ([xshift=\pgflinewidth]\pgf@node@name.south)  arc
                           (0: 0:\insiderad-\pgflinewidth)--cycle;            
         }}}}}  
 \makeatother

\graphicspath{ {images/} }

\begin{document}


\title{Detect, anticipate and generate: Semi-supervised recurrent latent variable models for human activity modeling}


\author
{\IEEEauthorblockN{Judith B\"utepage$^1$ }
\IEEEauthorblockA{ 
 butepage@kth.se}
\and
\IEEEauthorblockN{Danica Kragic$^1$}
\IEEEauthorblockA{
 dani@kth.se}
}
\maketitle


\begin{abstract}
Successful Human-Robot collaboration requires a predictive model of human behavior. The robot needs to be able to recognize current goals and actions and to predict future activities in a given context. However, the spatio-temporal sequence of human actions is difficult to model since latent factors such as intention, task, knowledge, intuition and preference determine the action choices of each individual. In this work we introduce semi-supervised variational recurrent neural networks which are able to a) model temporal distributions over latent factors and the observable feature space, b) incorporate discrete labels such as activity type when available, and c) generate possible future action sequences on both feature and label level. We evaluate our model on the Cornell Activity Dataset CAD-120 dataset. Our model outperforms state-of-the-art approaches in both activity and affordance detection and anticipation. Additionally, we show how samples of possible future action sequences are in line with past observations.    
\end{abstract}

\begin {IEEEkeywords}
Human behavior modeling, activity anticipation
\end{IEEEkeywords}

\let\thefootnote\relax\footnotetext{$^1$The Authors are with the Robotics, Perception and Learning Lab, EECS, KTH Royal Institute of Technology, Stockholm, Sweden. This work was supported by the EU through the project socSMCs (H2020-FETPROACT-2014) and the Swedish Foundation for Strategic Research. 
}

\section{Introduction}
\label{intro}
Human behavior is often stochastic and therefore difficult to predict over a longer period of time. Even within the context of a given task and a certain environment individuals might act differently based on e.g. intuition, prior knowledge and preferences. For example, if you provide a number of individuals with the task to prepare a meal following the same recipe, one person might follow a different order than specified because they have learned that a certain ingredient needs time to develop flavor. Another person might use only the big green knife instead of the more handy red knife because they prefer the color green and someone else might intentionally leave out a step.   

One way to approach this problem is to model different types of human characters \cite{nikolaidis2015efficient}. While this method is suitable for a single task setting such as an assembly line application, it might not scale to more general behavior which is distributed over many tasks and environments.
A more scalable approach is structured prediction with e.g. conditional random fields (CRFs) \cite{koppula2013learning,koppula2016anticipating1} which allows to capture the statistical dependencies between human subjects, their activities, objects in the environment and their affordances. However, common CRFs are limited in their capacity to model long-term dependencies due to the Markov assumption. 
Structural recurrent neural networks (S-RNN) \cite{jain2015structural} overcome this problem by employing recurrent neural networks (RNNs) as nodes and edges in the structured graph to detect and predict activity and affordance labels at each time step. The expressiveness and representational power of these neural networks increases the predictive power over short time horizons but the model structure prohibits long-term sequence generation. As S-RNNs do not explicitly learn to predict future feature states, they can not generate possible state-action sequences. Additionally, this deterministic model is not able to generate multiple possible sequences but is restricted to predict a single label. 

The key contribution of this paper is to address these issues with a generative, temporal model that can capture the complex dependencies of context and human features as well as discrete, hierarchical labels over time.
In detail, we propose a semi-supervised variational recurrent neural network (SVRNN), as described in Section \ref{sec:SVRNN}, which inherits the generative capacities of a variational autoencoder (VAE) \cite{kingma2013auto, rezende2014stochastic}, extends these to temporal data \cite{chung2015recurrent} and combines them with a discriminative model in a semi-supervised fashion. The semi-supervised VAE, first introduced by \cite{kingma2014semi}, can handle labeled and unlabeled data. This property allows us to propagate label information over time even during testing and therefore to generate possible future action sequences. 
Furthermore, we incorporate the dependencies between human and object features by extending the model to a multi-entity semi-supervised variational recurrent neural network (ME-SVRNN), as introduced in Section  \ref{sec:meSVRNN}.  The ME-SVRNN propagates information about the current state of an entity to other entities which increases the predictive power of the model. 
We apply our model to the Cornell Activity Dataset (CAD-120), consisting of 4 subjects who perform ten different high level actions, see Section \ref{section3} for details. Our model is trained to simultaneously detect and anticipate the activities and object affordances and to predict the next time step in feature space. We find that our model outperforms state-of-the-art methods in both detection and anticipation (Section \ref{sec:detection}) while being able to generate possible long term action sequences (Section \ref{sec:generation}). We conclude this paper with a final discussion of these findings in Section \ref{section4}.

\section{Methodology}
\label{section2}
In this section we introduce the model structure and detail the inference procedure. After a short overview of VAEs, we begin with a description of the general SVRNN before extending it to the multi-entity case. 

We denote random variables by bold characters and represent continuous data points by $\mathbf{x}$, discrete labels by  $\mathbf{y}$ and latent variables by  $\mathbf{z}$. The hidden state of a RNN unit at time $t$ is denoted by $h_t$. Similarly, time-dependent random variables are indexed by $t$, e.g. $\mathbf{x_t}$. Distributions $p_\theta$ commonly depend on parameters $\theta$. For the sake of brevity, we will neglect this dependence in the following discussion.

 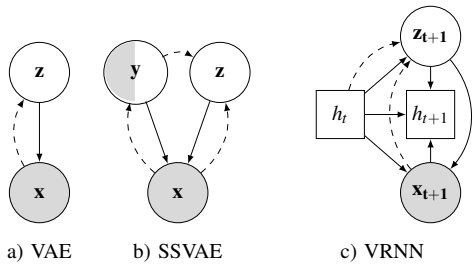
\begin{figure}[b!]
 \vspace{-0.3cm}
\centering   
\scalebox{0.8}{\pgfdeclarelayer{background}
\pgfdeclarelayer{foreground}
\pgfsetlayers{background,main,foreground}

\begin{tikzpicture}

\centering

\tikzstyle{observed} = [circle,draw=black, fill=gray!30,  minimum size=1cm,  inner sep=1.5pt]
\tikzstyle{unobserved} = [circle,draw=black, fill=white!30,  minimum size=1cm,  inner sep=1.5pt]

\tikzstyle{semiobserved} = [circle,draw=black, fill=white!30,pattern=north east lines, pattern color=gray,  minimum size=1cm,  inner sep=1.5pt]

\tikzstyle{empty} = [circle,draw=black, fill=white!0,minimum size=1.06cm,  inner sep=1.5pt]

\tikzstyle{rnnnode} = [rectangle,draw=black, fill=white!30,  minimum size=0.8cm,  inner sep=1.5pt]

\tikzstyle{textnode} = [rectangle,draw=white, fill=white!30,  minimum size=0.8cm,  inner sep=1.5pt]

\tikzstyle{condition} = [ellipse,draw=black, fill=green!30,    inner sep=2.5pt]
\tikzstyle{action} = [rectangle,draw=black, fill=yellow!30,   inner sep=2.5pt]
\tikzstyle{control} = [regular polygon,regular polygon sides=4, draw, fill=white!11,  text badly centered,  text width=1.0em,  inner sep=1.5pt]
\tikzstyle{arrowline} = [draw,color=black, -latex]
\tikzstyle{arrowlined} = [draw,  color=black, -latex , out=5  ]
\tikzstyle{arrowdashed} = [draw,dashed, color=black, -latex  ]
\tikzstyle{arrowdash} = [draw,dashed, color=black, -latex,  bend left=40]
\tikzstyle{arrowbend} = [draw, color=black, -latex, bend right=40]
\tikzstyle{arrowbendleft} = [draw, color=black, -latex, bend left=40]

\tikzstyle{arrowdashbend} = [draw,dashed, color= black, -latex, bend right=15]

\tikzstyle{arrowdashbendleft} = [draw,dashed, color= black, -latex, bend left=30]
\tikzstyle{arrowdashbendleftmore} = [draw,dashed, color= black, -latex, bend left=40]

\tikzstyle{arrowdashbendred} = [draw, dashed,  color= black,, -latex, bend right=35]
\tikzstyle{arrowdashbendless} = [draw, dashed,  color= black,, -latex, bend right=30 ]
 
\tikzstyle{textit} = [draw=none,fill=none]


\node [observed] at (2, 0.3) (nXt) {$\mathbf{x}$};
\node [unobserved] at (2, 2.3) (nZt) {$\mathbf{z}$};
\node [textnode] at (2, -0.7) (texta) {a) VAE};

\path [arrowline] (nZt) to (nXt); 
\path [arrowdashbendleft] (nXt) to (nZt);

\node [observed] at (4.3, 0.3) (Xt) {$\mathbf{x}$};
\node [unobserved] at (5, 2.3) (Zt) {$\mathbf{z}$};
\node [empty] at (3.6, 2.3) (ds) { };
 \node[shape=circle,
    draw=white ,  minimum size=1cm,  inner sep=1.5pt,
    circle split part fill={gray!30,white!30}
    ] at (3.6, 2.3) (Yt){$\mathbf{y}$};

empty

\node [textnode] at (4.3, -0.7) (textb) {b) SSVAE};
\path [arrowline] (Zt) to (Xt); 
\path [arrowdashbendless] (Xt) to (Zt);

\path [arrowline] (Yt) to (Xt); 
\path [arrowdashbendleft] (Xt) to (Yt);
\path [arrowdashbendleft] (Yt) to (Zt);


 \node [observed] at (8.5, 0.3) (nnXt1) {$\mathbf{x_{t+1}}$};

 \node [rnnnode] at ( 7,  1.6) (nnhzt) {$h_{t}$};

 \node [rnnnode] at ( 8.5,  1.6) (nnhzt1) {$h_{t+1}$};

\node [unobserved] at ( 8.5, 2.9) (nnCt1) {$\mathbf{z_{t+1}}$};

\node [textnode] at (7.7, -0.7) (textc) {c) VRNN};

 \path [arrowline] (nnhzt) to (nnhzt1); 
 \path [arrowline] (nnhzt) to (nnCt1); 
 \path [arrowline] (nnhzt) to (nnXt1);
 \path [arrowline] (nnCt1) to (nnhzt1);
 \path [arrowline] (nnXt1) to (nnhzt1);
 \path [arrowbendleft] (nnCt1) to (nnXt1);
 \path [arrowdashbendleftmore] (nnXt1) to (nnCt1);
 \path [arrowdashbendleft] (nnhzt) to (nnCt1);

\end{tikzpicture}}
\caption{Model structure of the VAE (a)), its semi-supervised version SVAE (b)), and the recurrent model VRNN (c)). Random variables (circle) and states of RNN hidden units (square) are either observed (gray), unobserved (white) or partially observed (gray-white). The dotted arrows indicate inference connections.} 
\label{fig:intromodels}
\end{figure}

\subsection{Variational autoencoders and amortized inference}
\label{sec:introtomethod}

Our model builds on VAEs, latent variable models that are combined with an amortized version of variational inference (VI). Amortized VI employs neural networks to learn a function from the data $\mathbf{x}$ to a distribution over the latent variables $q(\mathbf{z}|\mathbf{x})$  that approximates the posterior $p(\mathbf{z}|\mathbf{x})$. Likewise, they learn the likelihood distribution as a function of the latent variables $p(\mathbf{x}|\mathbf{z})$. This mapping is depicted in Figure \ref{fig:intromodels}a). Instead of having to infer $N$ local latent variables for $N$ observed data points, as common in VI, amortized VI requires only the learning of neural network parameters of the functions $q(\mathbf{z}|\mathbf{x})$ and $p(\mathbf{z}|\mathbf{x})$. We call $q(\mathbf{z}|\mathbf{x})$  the recognition network and  $p(\mathbf{z}|\mathbf{x})$ the generative network. 
To sample from a VAE, we first draw a sample from the prior $\mathbf{z} \sim p(\mathbf{z})$ which is then fed to the generative network to yield $\mathbf{x} \sim p(\mathbf{x}|\mathbf{z})$. We refer to \cite{zhang2017advances} for more details. 
 
To incorporate label information when available, semi-supervised VAEs (SVAE) \cite{kingma2014semi} include a label $\mathbf{y}$ into the generative process $p(\mathbf{x}|\mathbf{z}, \mathbf{y})$ and the recognition network  $q(\mathbf{z}|\mathbf{x}, \mathbf{y})$, as shown in Figure \ref{fig:intromodels}b). To handle unobserved labels, an additional approximate distribution over labels $q(\mathbf{y}|\mathbf{x})$ is learned which can be interpreted as a classifier. When no label is available, the discrete label distribution can be marginalize out, e.g. $q(\mathbf{z}|\mathbf{x}) = \sum_\mathbf{y}q(\mathbf{z}|\mathbf{x}, \mathbf{y})q(\mathbf{y}|\mathbf{x})$. 

VAEs can also be extended to temporal data, so called variational recurrent neural networks (VRNN)  \cite{chung2015recurrent}. Instead of being stationary as in vanilla VAEs, the prior over the latent variables depends in this case on past observations $p(\mathbf{z_t}|h_{t-1})$, which are encoded in the hidden state of a RNN $h_{t-1}$. Similarly, the approximate distribution $q(\mathbf{z_t}|\mathbf{x_t}, h_{t-1})$ depends on the history as can be seen in Figure \ref{fig:intromodels}c). The advantage of this structure is that data sequences can be generated by sampling from the temporal prior instead of an uninformed prior, i.e. $\mathbf{z_t} \sim p(\mathbf{z_t}|h_{t-1})$. 

\subsection{Semi-supervised variational recurrent neural network}
\label{sec:SVRNN}

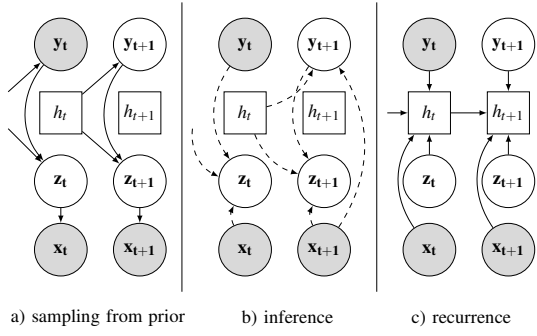
\begin{figure}[t!]
\centering   
\scalebox{0.7}{\pgfdeclarelayer{background}
\pgfdeclarelayer{foreground}
\pgfsetlayers{background,main,foreground}

\begin{tikzpicture}

\centering

\tikzstyle{observed} = [circle,draw=black, fill=gray!30,  minimum size=1cm,  inner sep=1.5pt]
\tikzstyle{unobserved} = [circle,draw=black, fill=white!30,  minimum size=1cm,  inner sep=1.5pt]

\tikzstyle{rnnnode} = [rectangle,draw=black, fill=white!30,  minimum size=0.8cm,  inner sep=1.5pt]

\tikzstyle{condition} = [ellipse,draw=black, fill=green!30,    inner sep=2.5pt]
\tikzstyle{action} = [rectangle,draw=black, fill=yellow!30,   inner sep=2.5pt]
\tikzstyle{control} = [regular polygon,regular polygon sides=4, draw, fill=white!11,  text badly centered,  text width=1.0em,  inner sep=1.5pt]
\tikzstyle{arrowline} = [draw,color=black, -latex]
\tikzstyle{arrowlined} = [draw,  color=black, -latex , out=5  ]
\tikzstyle{arrowdashed} = [draw,dashed, color=black, -latex , out=5  ]
\tikzstyle{arrowdash} = [draw,dashed, color=black, -latex,  bend left=40]
\tikzstyle{arrowbend} = [draw, color=black, -latex, bend right=40]
\tikzstyle{arrowbendleft} = [draw, color=black, -latex, bend left=35]

\tikzstyle{arrowdashbend} = [draw,dashed, color= black, -latex, bend right=25]

\tikzstyle{arrowdashbendleft} = [draw,dashed, color= black, -latex, bend left=25]

\tikzstyle{arrowdashbendred} = [draw, dashed,  color= black,, -latex, bend right=35]
\tikzstyle{arrowdashbendless} = [draw, dashed,  color= black,, -latex, bend right=33 ]
 
\tikzstyle{textit} = [draw=none,fill=none]

\node [observed] at (-4, 0.3) (Xt) {$\mathbf{x_t}$};
\node [observed] at (-2.5, 0.3) (Xt1) {$\mathbf{x_{t+1}}$};

\node [unobserved] at (-4, 1.6) (Zt) {$\mathbf{z_t}$};
\node [unobserved] at (-2.5, 1.6) (Zt1) {$\mathbf{z_{t+1}}$};

\node [rnnnode] at (-4, 2.9) (hzt) {$h_t$};
\node [rnnnode] at (-2.5, 2.9) (hzt1) {$h_{t+1}$};

\node [observed] at (-4, 4.2) (Ct) {$\mathbf{y_t}$};
\node [unobserved] at (-2.5, 4.2) (Ct1) {$\mathbf{y_{t+1}}$};

\node [textit] at (-3.3, -1) (a) {a) sampling from prior};

\path [arrowline] (hzt) to (Zt1); 

 \path [arrowline] (hzt) to (Ct1);

\path [arrowline] (-5,2.6) to (Zt);
\path [arrowline] (-5,3.3) to (Ct);

\path [arrowline] (Zt) to (Xt); 
\path [arrowline] (Zt1) to (Xt1); 
 
\path [arrowbend] (Ct) to (Zt);
\path [arrowbend] (Ct1) to (Zt1);



\node [observed] at (-0.5, 0.3) (nXt) {$\mathbf{x_t}$};
\node [observed] at (1.0, 0.3) (nXt1) {$\mathbf{x_{t+1}}$};

\node [unobserved] at (-0.5, 1.6) (nZt) {$\mathbf{z_t}$};
\node [unobserved] at ( 1, 1.6) (nZt1) {$\mathbf{z_{t+1}}$};

\node [rnnnode] at (-0.5, 2.9) (nhzt) {$h_t$};
\node [rnnnode] at ( 1, 2.9) (nhzt1) {$h_{t+1}$};

\node [observed] at (-0.5, 4.2) (nCt) {$\mathbf{y_t}$};
\node [unobserved] at ( 1, 4.2) (nCt1) {$\mathbf{y_{t+1}}$};

\path [arrowdashbendred] (nCt) to (nZt);  
\path [arrowdashbendred] (nCt1) to (nZt1);

\path [arrowdashbend] (nhzt) to (nZt1); 
\path [arrowdashbend] (nhzt) to (nCt1);

\path [arrowdashbendleft] (nXt) to (nZt);
\path [arrowdashbendleft] (nXt1) to (nZt1);

\path [arrowdashbendless] (nXt1) to (nCt1);

\path [arrowdashbendred] (-1.5,2.6) to (nZt);
 
  \draw (-1.7,-0.5) -- (-1.7,5);
\node [textit] at (0.3, -1) (b) {b) inference};


\node [observed] at (3, 0.3) (nnXt) {$\mathbf{x_t}$};
\node [observed] at (4.5, 0.3) (nnXt1) {$\mathbf{x_{t+1}}$};

\node [unobserved] at (3, 1.6) (nnZt) {$\mathbf{z_t}$};
\node [unobserved] at ( 4.5, 1.6) (nnZt1) {$\mathbf{z_{t+1}}$};

\node [rnnnode] at (3, 2.9) (nnhzt) {$h_t$};
\node [rnnnode] at ( 4.5, 2.9) (nnhzt1) {$h_{t+1}$};

\node [observed] at (3, 4.2) (nnCt) {$\mathbf{y_t}$};
\node [unobserved] at ( 4.5, 4.2) (nnCt1) {$\mathbf{y_{t+1}}$};

 \draw (2,-0.5) -- (2,5);

\path [arrowline] (nnhzt) to (nnhzt1);

\path [arrowline] (nnCt) to (nnhzt); 
\path [arrowline] (nnCt1) to (nnhzt1);  

\path [arrowline] (nnZt) to (nnhzt); 
\path [arrowline] (nnZt1) to (nnhzt1);  

\path [arrowline] (2.2 ,2.9) to (nnhzt); 
\path [arrowbendleft] (nnXt) to (nnhzt); 
\path [arrowbendleft] (nnXt1) to (nnhzt1);

 \node [textit] at (3.6, -1) (c) {c) recurrence};

\end{tikzpicture}}
\caption{Information flow through SVRNN. a) Passing samples from the prior through the generative network. b) Information passing through the inference network. c) The recurrent update. Node appearance follows Figure \ref{fig:intromodels}.} 
\label{fig:model}
\vspace{-0.5cm}
\end{figure}

For SVRNN, we assume that we are given a dataset with temporal structure $D = \{D^L, D^U\}$ consisting of $L$ labeled time steps $D^L = \{\bmxt,\bmyt\}_{t \in L} \sim \pem(\bmxt,\bmyt)$ and $U$ unlabeled observations $D^U = \{\bmxt\}_{t \in U} \sim  \pem(\bmxt)$. $\pem$ denotes the empirical distribution. Further we assume that the temporal process is governed by latent variables $\bmzt$, whose distribution $p(\mathbf{z_t}|h_{t-1})$ depends on a deterministic function of the history up to time $t$: $h_{t-1} = f(x_{<t}, y_{<t}, z_{<t})$. The generative process follows
$\mathbf{y_t} \sim p(\mathbf{y_t}|h_{t-1}), \ \mathbf{z_t} \sim p(\mathbf{z_t}|\mathbf{y_t}, h_{t-1})$ and finally $\mathbf{x_t} \sim p(\mathbf{x_t}|\mathbf{y_t}, \mathbf{z_t}, h_{t-1}).$
Here, $p(\mathbf{y_t}|h_{t-1})$ and $p(\mathbf{z_t}|\mathbf{y_t}, h_{t-1})$ are time-dependent priors, as shown in Figure \ref{fig:model}a). To fit this model to the dataset at hand, we need to estimate the posterior over the unobserved variables $p(\mathbf{y_t}|\mathbf{x_t}, h_{t-1})$ and $p(\mathbf{z_t}|\mathbf{x_t}, \mathbf{y_t}, h_{t-1})$ which is intractable. Therefore we resign to amortized VI and approximate the posterior with a simpler distribution  $q(\mathbf{y_t}, \mathbf{z_t}|\mathbf{x_t}, h_{t-1}) = q(\mathbf{y_t}|\mathbf{x_t}, h_{t-1})q(\mathbf{z_t}|\mathbf{x_t}, \mathbf{y_t}, h_{t-1})$, as shown in Figure \ref{fig:model}b). To minimize the distance between the approximate and posterior distributions, we optimize the variational lower bound of the marginal likelihood $\mathcal{L}(p(D)) $. As the distribution over $\mathbf{y_t}$ is only required when it is unobserved, the bound decomposes as follows
\begin{align}
& \quad \quad \quad \mathcal{L}  (p(D))  \geq  \mathcal{L}^L  + \mathcal{L}^U  + \alpha\mathcal{T}^L\label{eq:loglikone}\\
-\mathcal{L}^L &=  \sum_{t \in L}  \mathbb{E}_{q(\bmzt|\bmxt, \bmyt, \bmhlt )}[log(p(\bmxt| \bmyt, \bmzt, \bmhlt))] -   \nonumber  \\
     K&L(q(\bmzt|\bmxt, \bmyt , \bmhlt )||p(\bmzt|  \bmyt, \bmhlt )) + log(p(\bmyt))  \nonumber \\
 \mathcal{T}^L &=-\sum_{t \in L}\mathbb{E}_{\pem(\bmyt, \bmxt)} log (p(\bmyt|\bmhlt )q(\bmyt|\bmxt, \bmhlt ))  \nonumber \\
 -\mathcal{L}^U &=  \sum_{t \in U}  \mathbb{E}_{q(\bmyt, \bmzt|\bmxt, \bmhlt )} \big{[} log(p(\bmxt| \bmyt, \bmzt, \bmhlt ))] \nonumber  \\
 &   \quad -   KL(q(\bmzt|\bmxt, \bmyt, \bmhlt  )||p(\bmzt| \bmyt, \bmhlt  ))\big{]}  \nonumber \\
 &   \quad  -   KL(q(\bmyt|\bmxt, \bmhlt )||p(\bmyt|\bmhlt )).   \nonumber
\end{align}
$\mathcal{L}^L$ and $\mathcal{L}^U$ are the lower bounds for labeled and unlabeled data points respectively, while $\mathcal{T}^L$ is an additional term that encourages $p(\bmyt|\bmhlt)$ and $q(\bmyt|\bmxt, \bmhlt)$ to follow the data distribution over $\mathbf{y_t}$. This lower bound is optimized jointly. We assume the latent variables $\mathbf{z_t}$ to be i.i.d Gaussian distributed. The categorical distribution  over $\mathbf{y_t}$ is determined by parameters $\pi = \{\pi_i\}_{i=1:N_class})$. To model such discrete distributions, we apply the Gumbel trick \cite{jang2016categorical,maddison2016concrete}. The history $h_{t-1}$ is modeled with a Long short-term memory (LSTM) unit \cite{hochreiter1997long}. For more details, we refer the reader to the related work discussed in Section \ref{sec:introtomethod}.

\subsection{Modeling multiple entities }
\label{sec:meSVRNN}


To model different entities, we allow these to share information between each other over time. The structure and information flow of this model is a design choice. In our case, these entities consist of the human $H$ and $o \in [1, N_o]$ additional entities, such as objects or other humans. We denote the dependency of variables on their source by $(\mathbf{x_t}^H, \mathbf{y_t}^H, \mathbf{z_t}^H, h_t^H)$ and $\{(\mathbf{x_t}^o, \mathbf{y_t}^o, \mathbf{z_t}^o, h_t^o)\}_{o \in 1:N_o}$. Further, we summarize the history and current observation of all additional entities by $h_t^O=\sum_{o}h_t^o$ and  $\mathbf{x_t}^O=\sum_{o}\mathbf{x_t}^o$ respectively. Instead of only conditioning on its own history and observation, as described in Section \ref{sec:SVRNN}, we let the entities share information by conditioning on others' history and observations. Specifically, the model of the human receives information from all additional entities, while these receive information from the human model. 
Let $\mathbf{x_t^{AB}} = [\mathbf{x^A_{t}}, \mathbf{x_{t}^B}]$ and $h_t^{AB} = [h^A_{t}, h_{t}^B]$ for $A,B \in (H,O,o)$.
The structure of the prior and approximate distribution then become  
$p(\mathbf{\mathbf{y^H_t}}|h^{HO}_{t-1})$, $p(\mathbf{\mathbf{z^H_t}}|\mathbf{y^H_t}, h^{HO}_{t-1})$,
$q(\mathbf{\mathbf{y^H_t}}|\mathbf{x^{HO}_t}, h_{t-1}^{HO})$ and $q(\mathbf{\mathbf{z^H_t}}|\mathbf{x^{HO}_t}, \mathbf{y^H_t}, h_{t-1}^{HO})$ for the human, and $p(\mathbf{\mathbf{y^o_t}}|h^{oH}_{t-1})$, $p(\mathbf{z^o_t}|\mathbf{y^o_t}, h^{oH}_{t-1})$, $q(\mathbf{\mathbf{y^o_t}}|\mathbf{x^{oH}_t}, h^{oH}_{t-1})$ and $q(\mathbf{z^o_t}| \mathbf{x^{oH}_t}, \mathbf{y^o_t}, h_{t-1}^{oH})$ for each additional entity $o \in  1:N_o$, We assume that the labels for all entities are observed and unobserved at the same points in time. Therefore, the lower bound in Equation \ref{eq:loglikone} is extended by summing over all entities.

\section{Experiments}
\label{section3}

\begin{figure*}[t]
\begin{minipage}[c]{0.65\textwidth}
    \includegraphics[width=\textwidth]{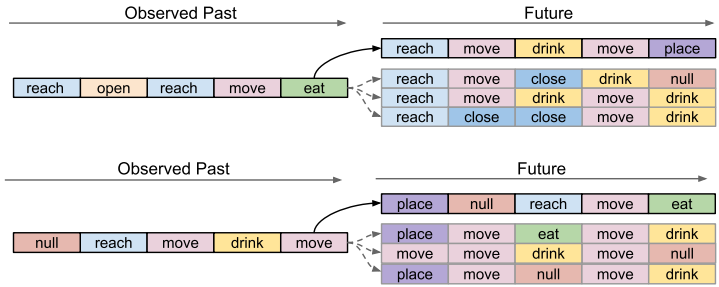}
  \end{minipage}\hfill
  \begin{minipage}[c]{0.35\textwidth}
    \caption{
      Sampled sub-activity sequences given the last five observed sub-activities of the high-level actions \textit{taking medicine} (top) and \textit{having a meal} (bottom). Black lines indicate ground truth and gray lines indicate sampled sub-activities. A sub-activity has an average duration of 3.6 seconds.
    } \label{fig:samples}
  \end{minipage}
\end{figure*}


In this section, we present our experimental results. 
We evaluate our model on the Cornell Activity Dataset 120 (CAD -120) \cite{koppula2013learning}. This dataset consists of 4 subjects performing 10 high-level tasks, such as \textit{cleaning a microwave} or \textit{having a meal}, in 3 trials each. These activities are further annotated with 10 sub-activities, such as \textit{moving} and \textit{eating} and 12 object affordances, such as \textit{movable} and \textit{openable}. In this work we are focusing on classifying the sub-activities and affordances. We use the features extracted in \cite{koppula2013learning} and preprocess these as in \cite{jain2015structural}. Our results rely on four-fold cross-validation with the same folds as used in \cite{koppula2013learning}. For comparison, we trained the S-RNN models, for which code is provided online, on these folds and under the same conditions as described in \cite{jain2015structural}. We use a learning rate of 0.001, a batch size of 10 and the adagrad optimizer. Further, we apply a dropout rate of 0.1 to all units but the latent variable parameters and the output layers. In each batch, we mark ca. 25 \% of the labels as unobserved. The object models share all parameters, i.e. we effectively learn one human model and one object model both in the single- and multi-entity case.

\subsection{Detection and anticipation}
\label{sec:detection}

First, we investigate the ability of our model to detect the current sub-activity and object affordance and to anticipate these labels at the following time step. We compare the performance to the anticipatory CRF reported in  \cite{koppula2016anticipating1} and the replicated results of the S-RNN \cite{jain2015structural}. The F1 score of all models averaged over the cross-validation folds and 20 samples from the latent distributions is reported in Table \ref{tab:F1score}. While the SVRNN without information exchange between entities outperforms the baseline methods, the multi-entity model achieves the highest values. Especially the sub-activity detection and anticipation seems to benefit from the information provided by the object states and observations. 

\begin{table}[h!]     
\begin{tabular}{|c|c|c|c|c|}
\hline
 & \multicolumn{2}{|c|}{Detection}  & \multicolumn{2}{|c|}{Anticipation}  \\
\hline
Method &  Sub-Act  & Obj-Aff & Sub-Act  & Obj-Aff\\ \hline
ATCRF \cite{koppula2016anticipating1}  &  86.4  & 85.2 &  40.6 & 41.4 \\ \hline
S-RNN \cite{jain2015structural} & 69.6  & 84.8 & 53.9 & 74.3 \\ \hline
SVRNN & 83.4  & 88.3 & 67.7 & 81.4\\ \hline
ME-SVRNN & \textbf{89.8}  &  \textbf{90.5} &  \textbf{77.1} &  \textbf{82.1} \\ \hline
\end{tabular}
\caption{\label{tab:F1score}Average F1 score for sub-activity and object affordances for detection and anticipation.}
\end{table}

\subsection{Generation}
\label{sec:generation}

In contrast to S-RNN, our SVRNN model is able to generate possible, long-term action sequences. These are generated by propagating a short observation sequence through the network to obtain the summarizing state $h_{t-1}$ and to subsequently sample from the priors $p(\bmzt|  \bmhlt )$ and $p(\bmyt|  \bmhlt )$. These samples are used by the generative network to make a prediction of the next observation $\mathbf{\hat{x}_t}$, which forms the next input to the model. We present a number of sampled sub-activity sequences in Figure \ref{fig:samples}. Note that a sub-activity has an average duration of 3.6 seconds \cite{koppula2016anticipating1}. Thus, we sample possible sequences for around 18 seconds into the future. The samples are plausible action sequences given the observed past. For example, the model remembers that the action \textit{opening} requires \textit{closing} over several time steps. Additionally, unrelated sub-activities such as \textit{cleaning} are not sampled.


\section{Conclusion}
\label{section4}
In this work, we presented a generative, temporal model for human activity modeling. Our experimental evaluation shows promising performance in the three tasks of detection, anticipation and generation. In future work, we are planning to evaluate the model more extensively and to extend the model to hierarchical label structures. 
   

\bibliographystyle{plain}
\bibliography{main}

\begin{thebibliography}{10}

\bibitem{chung2015recurrent}
Junyoung Chung, Kyle Kastner, Laurent Dinh, Kratarth Goel, Aaron~C Courville,
  and Yoshua Bengio.
\newblock A recurrent latent variable model for sequential data.
\newblock In {\em NIPS}, pages 2980--2988, 2015.

\bibitem{hochreiter1997long}
Sepp Hochreiter and J{\"u}rgen Schmidhuber.
\newblock Long short-term memory.
\newblock {\em Neural computation}, 9(8):1735--1780, 1997.

\bibitem{jain2015structural}
Ashesh Jain, Amir~R Zamir, Silvio Savarese, and Ashutosh Saxena.
\newblock {Structural-RNN}: Deep learning on spatio-temporal graphs.
\newblock In {\em IEEE Conference on Computer Vision and Pattern Recognition},
  2016.

\bibitem{jang2016categorical}
Eric Jang, Shixiang Gu, and Ben Poole.
\newblock Categorical reparameterization with gumbel-softmax.
\newblock In {\em ICLR}, 2017.

\bibitem{kingma2014semi}
Diederik~P Kingma, Shakir Mohamed, Danilo~Jimenez Rezende, and Max Welling.
\newblock Semi-supervised learning with deep generative models.
\newblock In {\em NIPS}, pages 3581--3589, 2014.

\bibitem{kingma2013auto}
Diederik~P Kingma and Max Welling.
\newblock Auto-encoding variational bayes.
\newblock {\em arXiv preprint arXiv:$1312.6114$}, 2013.

\bibitem{koppula2016anticipating1}
Hema~S Koppula and Ashutosh Saxena.
\newblock Anticipating human activities using object affordances for reactive
  robotic response.
\newblock {\em IEEE transactions on pattern analysis and machine intelligence},
  38(1):14--29, 2016.

\bibitem{koppula2013learning}
Hema~Swetha Koppula, Rudhir Gupta, and Ashutosh Saxena.
\newblock Learning human activities and object affordances from rgb-d videos.
\newblock {\em The International Journal of Robotics Research}, 32(8):951--970,
  2013.

\bibitem{maddison2016concrete}
Chris~J Maddison, Andriy Mnih, and Yee~Whye Teh.
\newblock The concrete distribution: A continuous relaxation of discrete random
  variables.
\newblock In {\em ICLR}, 2017.

\bibitem{nikolaidis2015efficient}
Stefanos Nikolaidis, Ramya Ramakrishnan, Keren Gu, and Julie Shah.
\newblock Efficient model learning from joint-action demonstrations for
  human-robot collaborative tasks.
\newblock In {\em 2015 10th ACM/IEEE International Conference on HRI}, pages
  189--196. ACM, 2015.

\bibitem{rezende2014stochastic}
Danilo~Jimenez Rezende, Shakir Mohamed, and Daan Wierstra.
\newblock Stochastic backpropagation and approximate inference in deep
  generative models.
\newblock {\em arXiv preprint arXiv:1401.4082}, 2014.

\bibitem{zhang2017advances}
Cheng Zhang, Judith Butepage, Hedvig Kjellstrom, and Stephan Mandt.
\newblock Advances in variational inference.
\newblock {\em arXiv preprint arXiv:1711.05597}, 2017.

\end{thebibliography}

\end{document}